# Fusing Color and Texture Cues to Categorize the Fruit Diseases from Images

Shiv Ram Dubey, Anand Singh Jalal

## ABSTRACT

*The economic and production losses in agricultural industry worldwide are due to the presence of diseases in the several kinds of fruits. In this paper, a method for the classification of fruit diseases is proposed and experimentally validated. The image processing based proposed approach is composed of the following main steps; in the first step K-Means clustering technique is used for the defect segmentation, in the second step color and textural cues are extracted and fused from the segmented image, and finally images are classified into one of the classes by using a Multi-class Support Vector Machine. We have considered diseases of apple as a test case and evaluated our approach for three types of apple diseases namely apple scab, apple blotch and apple rot and normal apples without diseases. Our experimentation points out that the proposed fusion scheme can significantly support accurate detection and automatic classification of fruit diseases.*

## INTRODUCTION

Recognition system is a 'grand challenge' for the computer vision to achieve near human levels of recognition. In the agricultural sciences, images are the important source of data and information. To reproduce and report such data photography was the only method used in recent years. It is difficult to process or quantify the photographic data mathematically. Digital image analysis and image processing technology circumvent these problems based on the advances in computers and microelectronics associated with traditional photography. This tool helps to improve images from microscopic to telescopic visual range and offers a scope for their analysis. Monitoring of health and detection of diseases is critical in fruits and trees for sustainable agriculture. To the best of our knowledge, no sensor is available commercially for the real time assessment of trees health conditions. Scouting is the most widely used method for monitoring stress in trees, but it is expensive, time-consuming and labor-intensive process. Polymerase chain reaction which is a molecular technique used for the identification of fruit diseases but it requires detailed sampling and processing.

The various types of diseases on fruits determine the quality, quantity, and stability of yield. The diseases in fruits not only reduce the yield but also deteriorate the variety and its withdrawal from the cultivation. Early detection of disease and crop health can facilitate the control of fruit diseases through proper management approaches such as vector control through fungicide applications, disease-specific chemical applications and pesticide applications; and improved productivity. The classical approach for detection and identification of fruit diseases is based on the naked eye observation by experts. In some of the developing countries, consultation with experts is a time consuming and costly affair due to the distant locations of their availability.

Fruit diseases can cause significant losses in quality and yield appeared at the time of harvesting. For example, soybean rust (a fungal disease in soybeans) has caused a significant economic loss and just by removing 20% of the infection, the farmers may benefit with an approximately 11 million-dollar profit (Roberts et al., 2006). Some fruit diseases also infect other areas of the tree causing diseases of twigs, leaves and branches.

An early detection of fruit diseases can aid in decreasing such losses and can stop further spread of diseases. A lot of work has been done to automate the visual inspection of the fruits by machine vision with respect to size and color. However, detection of defects in the fruits using images is still problematic due to the natural variability of skin color in different types of fruits, high variance of defect types, and presence of stem/calyx. To know what control factors to consider next year to overcome similar losses, it is of great significance to analyze what is being observed. Some common diseases of apple fruits are apple scab, apple rot, and apple blotch (Hartman, 2010). Apple scabs are gray or brown corky spots. Apple rot infections produce slightly sunken, circular brown or black spots that may be covered by a red halo. Apple blotch is a fungal disease and appears on the surface of the fruit as dark, irregular or lobed edges.

In this paper, we introduce and experimentally evaluate a method for the classification of fruit diseases from images. The proposed method is composed of the following steps; in first step the infected part of fruit images are detected and segmented using K-Means clustering technique, in second step, some state-of-the-art color and texture features are extracted from the segmented image and fused to achieve the more discriminative characteristics, and finally, fruit diseases are classified using a Multi-class Support Vector Machine. We show the significance of using clustering technique for the disease segmentation and Multi-class Support Vector Machine as a classifier for the automatic classification of fruit diseases. In order to validate the proposed method, we have considered three types of the diseases in apple; apple blotch, apple rot and apple scab and also normal apples as test suite. The experimental results show that the proposed method can significantly achieve automatic detection and accurate classification of apple fruit diseases.

## LITERATURE REVIEW

In this section, we focus on the previous work done by several researchers in the area of image categorization, fruit and vegetable classification and fruit diseases identification. Fruit disease identification can be seen as an instance of image categorization. A framework for fruits and vegetables recognition and classification is proposed (Dubey, & Jalal, 2012a; Dubey, & Jalal, 2013; Dubey, 2012). They have considered images of 15 different types of fruit and vegetable collected from a supermarket. Their approach first segment the image to extract the region of interest and then calculate image features from that segmented region which is further used in training and classification by a multi-class support vector machine.

Recently, a lot of activity in the area of defect detection has been done. Major works performing defect segmentation of fruits are done using simple threshold approach (Li, Wang, & Gu, 2002). A globally adaptive threshold method (modified version of Otsu's approach) to segment fecal contamination defects on apples are presented by Kim et al. (2005). Classification-based methods attempt to partition pixels into different classes using different classification methods. Bayesian classification is the most used method (Kleynen, Leemans, & Destain, 2005) where pixels are compared with a pre-calculated model and classified as defected or healthy. Unsupervised classification does not benefit any guidance in the learning process due to lack of target values. This type of approach was used by Leemans, Magein, & Destain (1998) for defect segmentation. K-mean clustering based defect segmentation approaches have shown very accurate detection result. In this paper also, we use K-means clustering approach for defect segmentation. The image features used in (Agrawal 2014; Semwal, Sati, & Verma, 2011) can also be in integrated with the fruit disease recognition to improve the efficiency of the approach.

The spectroscopic and imaging techniques are unique disease monitoring approaches that have been used to detect diseases and stress due to various factors, in plants and trees. Current research activities are towards the development of such technologies to create a practical tool for a large-scale real-time disease monitoring under field conditions. Various spectroscopic and imaging techniques have been studied for the detection of symptomatic and asymptomatic plant and fruit diseases. Some the methods are: fluorescence imaging (Bravo et al., 2004); multispectral or hyperspectral imaging (Moshou et al., 2006); infrared spectroscopy (Spinelli,

Noferini, & Costa, 2006); visible/multiband spectroscopy (Yang, Cheng, & Chen, 2007), and nuclear magnetic resonance (NMR) spectroscopy (Choi et al., 2004). Hahn (2009) reviewed multiple methods (sensors and algorithms) for pathogen detection, with special emphasis on postharvest diseases. Several techniques for detecting plant diseases is reviewed by Sankarana et al. (2010) such as, Molecular techniques, Spectroscopic techniques (Fluorescence spectroscopy and Visible and infrared spectroscopy), and Imaging techniques (Fluorescence imaging and Hyper-spectral imaging).

Large scale plantation of oil palm trees requires on-time detection of diseases as the ganoderma basal stem rot disease was present in more than 50% of the oil palm plantations in Peninsular Malaysia. To deal with this problem, airborne hyper-spectral imagery offers a better solution (Shafri & Hamdan, 2009) in order to detect and map the oil palm trees that were affected by the disease on time. Airborne hyper-spectral has provided data on user requirement and has the capability of acquiring data in narrow and contiguous spectral bands which makes it possible to discriminate between healthy and diseased plants better compared to multispectral imagery. In (Qin et al., 2009), a hyper-spectral imaging approach is developed for detecting canker lesions on citrus fruit and hyper-spectral imaging system is developed for acquiring reflectance images from citrus samples in the spectral region from 450 to 930 nm.

Fernando et al. (2010) used an unsupervised method based on a Multivariate Image Analysis strategy which uses Principal Component Analysis (PCA) to generate a reference eigenspace from a matrix obtained by unfolding spatial and color data from defect-free peel samples. In addition, a multiresolution concept is introduced to speed up the process. They tested on 120 samples of mandarins and oranges from four different cultivars: Marisol, Fortune, Clemenules, and Valencia. They reported 91.5% success ratio for individual defect detection, while 94.2% classification ratio for damaged/sound samples. Fruit diseases are also recognized using image processing techniques (Dubey, & Jalal, 2012b). First of all, they detected the defected region by k-means clustering based image segmentation technique then extracted the features from that segmented defected region which is used by a multi-class support vector machine for training and classification purpose.

Gabriel et al. (2013) proposed a pattern recognition method to automatically detect stem and calyx ends and damaged blueberries. First, color and geometrical features were extracted. Second, five algorithms were tested to select the best features. The best classifiers were and Linear Discriminant Analysis. Using Support Vector Machine classifier, they distinguished the blueberries' orientation in 96.8% of the cases. The average performance for mechanically damaged, shriveled, and fungally decayed blueberries were reported as 86%, 93.3%, and 97% respectively. A synthesis segmentation algorithm is developed for the real-time online diseased strawberry images in greenhouse (Ouyang et al., 2013). The impact of uneven illumination is eliminated through the "top-hat" transform, and noise interferences are removed by median filtering. They obtained complete strawberry fruit area of the image after applying the methods of gray morphology, logical operation, OTSU and mean shift segmentation. Then, they normalize the extracted eigenvalues, and used eigenvectors of samples for training the support vector machine and BP neural network. Their Results indicate that support vector machines have higher recognition accuracy than the BP neural network.

Figure 1: Fruit disease recognition system

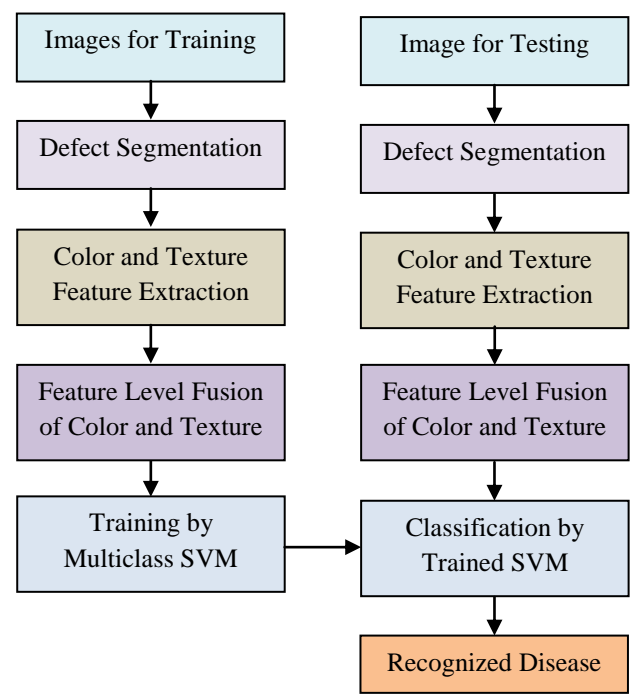

## FRUIT DISEASE CLASSIFICATION

Image categorization, in general, relies on combinations of structural, statistical and spectral approaches. Structural approaches describe the appearance of the object using well-known primitives, for example, patches of important parts of the object.

Statistical approaches represent the objects using local and global descriptors such as mean, variance, and entropy. Finally, spectral approaches use some spectral space representation to describe the objects such as Fourier spectrum (Gonzalez & Woods, 2007). In this paper, we introduce a method which exploits statistical color and texture descriptors to identify fruit diseases in a multi-class scenario.

The steps of the proposed approach are defect segmentation; feature extraction; training and classification as shown in Figure 1. For the fruit disease identification problem, precise image segmentation is required; otherwise the features of the non-infected region will dominate over the features of the infected region. K-means based defect segmentation is used to detect the region of interest which is the infected part only in the image. We extract the color and texture features from the segmented portion of the image and fuse them to obtain more distinctive description. We train support vector machine with the features stored in the feature database because it is required to learn the system with the characteristics of each type of diseases. Finally any input image can be classified into one of the classes using feature derived from segmented part of the input image and trained support vector machine.

## Defect Segmentation

In this paper, K-means clustering technique is used for the defect segmentation similar to Dubey and Jalal (2014a). Images are partitioned into four clusters in which one or more cluster contains only infected region of the fruit. The K-means clustering algorithms classify the objects (pixels in our problem) into K number of classes based on a set of features. The classification is carried out by minimizing the sum of squares of distances between the data objects and the corresponding cluster.

*Algorithm for the K-Means image segmentation –*
1. Read input image.
2. Transform image from RGB to L*a*b* color space.
3. Classify colors using K-Means clustering in 'a*b*' space.
4. Label each pixel in the image from the results of K-means.
5. Generate images that segment the image by color.
6. Select the segment containing disease.

In this experiment, squared Euclidean distance is used for the K-means clustering. We use L*a*b* color space because the color information in the L*a*b* color space is stored in only two channels (i.e. a* and b* components), and it causes reduced processing time for the defect segmentation. In this experiment input images are partitioned into four segments. From the empirical observations it is found that using 3 or 4 cluster yields good segmentation results. Figure 2 depicts some defect segmentation results using the K-mean clustering technique.

## Feature Extraction

We have used some state-of-the-art color and texture features and fused them to validate the accuracy and efficiency of the proposed approach. The features used for the fruit disease classification problem are Global Color Histogram, Color Coherence Vector, Color Difference Histogram, Structure Element Histogram, Local Binary Pattern, Local Ternary Pattern, and Completed Local Binary Pattern.

*Global Color Histogram (GCH)*

The Global Color Histogram (GCH) is the simplest approach to encode the information present in an image (Gonzalez & Woods, 2007).

Figure 2: Some defect segmentation results (a) Images before segmentation, (b) Images after segmentation

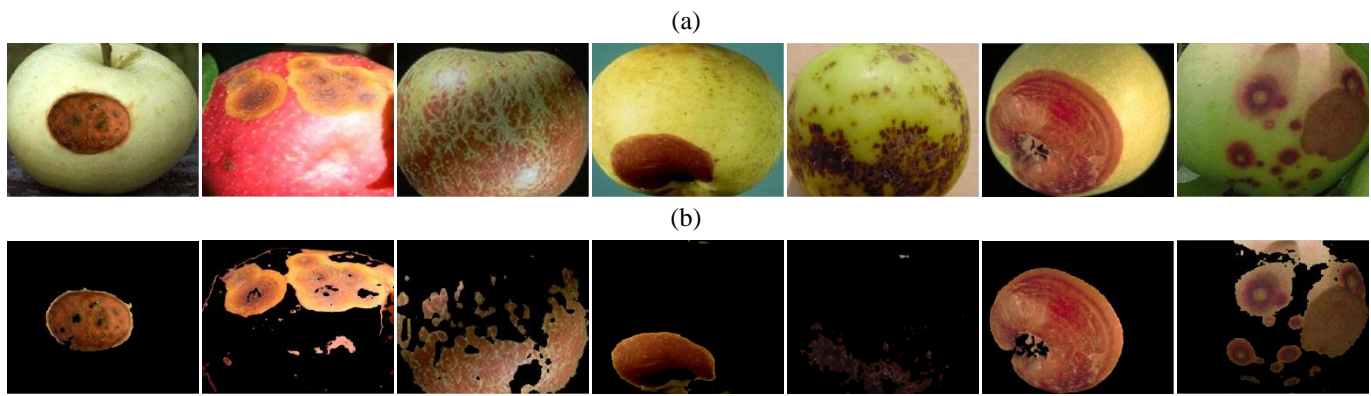

(a)

(b)

A GCH is a set of ordered values, for each distinct color, representing the probability of a pixel being of that color. Uniform normalization and quantization are used to avoid scaling bias and to reduce the number of distinct colors (Gonzalez & Woods, 2007).

*Color Coherence Vector (CCV)*

An approach to compare images based on color coherence vectors are presented by Pass, Zabih, & Miller (1997). They define color coherence as the degree to which image pixels of that color are members of a large region with homogeneous color. These regions are referred as coherent regions. Coherent pixels are belongs to some sizable contiguous region, whereas incoherent pixels are not. In order to compute the CCVs, the method blurs and discretizes the image's color-space to eliminate small variations between neighboring pixels. Then, it finds the connected components in the image in order to classify the pixels of a given color bucket is either coherent or incoherent. After classifying the image pixels, CCV computes two color histograms: one for coherent pixels and another for incoherent pixels. The two histograms are stored as a single histogram.

*Color Difference Histogram (CDH)*

A feature descriptor called as color difference histograms (CDH) is designed by using color differences of neighboring pixels at a certain distance (Liu, & Yang, 2013). The unique characteristic of CDH is that it count the perceptually uniform color difference between two points under different backgrounds with regard to colors and edge orientations in L*a*b* color space. It pays more attention to color, edge orientation and perceptually uniform color differences, and encodes color, orientation and perceptually uniform color difference via feature representation in a similar manner to the human visual system.

*Structure Element Histogram (SEH)*

A texture descriptor called structure elements' histogram (SEH) is proposed to encode the small local structures of the image (Xingyuan, & Zongyu, 2013). SEH describes images with its local features. It uses HSV color space (it has been quantized to 72 bins). SEH integrates the advantages of both statistical and structural texture description methods, and it can represent the spatial correlation of local textures.

*Local Binary Pattern (LBP)*

Given a pixel in the input image, LBP is computed by comparing it with its neighbors (Ojala, Pietikäinen, & Mäenpää, 2002):

$$LBP_{N,R} = \sum_{n=0}^{n-1} s(v_n - v_c)2^n, s(x) = \begin{cases} 1, x \geq 0 \\ 0, x < 0 \end{cases} \quad (1)$$

Where, $v_c$ is the value of the central pixel, $v_n$ is the value of its neighbors, $R$ is the radius of the neighborhood and $N$ is the total number of neighbors. Suppose the coordinate of $v_c$ is (0, 0), then the coordinates of $v_n$ are $(R\cos(2\pi n/N), R\sin(2\pi n/N))$. The values of neighbors that are not present in the image grids may be estimated by interpolation. Let the size of image is $I*J$. After the LBP code of each pixel is computed, a histogram is created to represent the texture image:

$$H(k) = \sum_{i=1}^{I} \sum_{j=1}^{J} f(LBP_{N,R}(i,j),k), k \in [0,K],$$
$$f(x,y) = \begin{cases} 1, x = y \\ 0, otherwise \end{cases} \quad (2)$$

Where, $K$ is the maximal LBP code value. In this experiment the value of '$N$' and '$R$' are set to '8' and '1' respectively to compute the LBP feature.

*Local Ternary Pattern (LTP)*

Local ternary pattern is a natural extension of the original LBP. In (Tan, & Triggs, 2010), Tan et al. proposed to use a base-3 pattern to represent the region. As a computationally efficient local image texture descriptor, LTP has been used with considerable success in a number of visual recognition tasks. The LTP can be calculated according to the following equation:

$$LTP(i) = \begin{cases} 1 & if\ P(i) - P(0) > \theta \\ 0 & if\ |P(i) - P(0)| \leq \theta \\ -1 & if\ P(i) - P(0) < -\theta \end{cases} \quad (3)$$

Where $P(0)$ is the intensity of the center pixel, and $\theta$ is a pre-defined threshold.

*Completed Local Binary Pattern (CLBP)*

LBP feature considers only signs of local differences (i.e. difference of each pixel with its neighbors) whereas CLBP feature considers both signs (S) and magnitude (M) of local differences as well as original center gray level (C) value (Guo, Zhang, & Zhang, 2010). CLBP feature is the combination of three features, namely CLBP_S, CLBP_M, and CLBP_C. CLBP_S is the same as the original LBP and used to code the sign information of local differences. CLBP_M is used to code the magnitude information of local differences:

$$CLBP_{N,R} = \sum_{n=0}^{n-1} t(m_n, c)2^n, t(x,c) = \begin{cases} 1, x \geq c \\ 0, x < c \end{cases} \quad (4)$$

Where, $c$ is a threshold and set to the mean value of the input image in this experiment.

CLBP_C is used to code the information of original center gray level value:

$$CLBP_{N,R} = t(g_c, c_I), t(x,c) = \begin{cases} 1, x \geq c \\ 0, x < c \end{cases} \quad (5)$$

Where, threshold $c_I$ is set to the average gray level of the input image. In this experiment the value of '$N$' and '$R$' are set to '8' and '1' respectively to compute the CLBP feature.

## Training and Classification using Multi-class Support Vector Machine

Supervised learning is a machine learning approach that aims to estimate a classification function f from a training data set. The trivial output of the function f is a label (class indicator) of the input object under analysis. The learning task is to predict the function outcome of any valid input object after having seen a sufficient number of training examples.

Recently, a unified approach is presented by Rocha et al. (2010) that can combine many features and classifiers. The author approaches the multi-class classification problem as a set of binary classification problem in such a way one can assemble together diverse features and classifier approaches custom-tailored to parts of the problem. They define a class binarization as a mapping of a multi-class problem onto two-class problems (divide-and-conquer) and referred binary classifier as a base learner. For N-class problem $N \times (N-1)/2$ binary classifiers will be needed where $N$ is the number of different classes. According to the author, the $ij^{th}$ binary classifier uses the patterns of class $i$ as positive and the patterns of class $j$ as negative. They calculate the minimum distance of the generated vector (binary outcomes) to the binary pattern (ID) representing each class, in order to find the final outcome. They have categorized the test case into a class for which distance between ID of that class and binary outcomes is minimum.

This approach can be understood by a simple three class problem. Let three classes are $x$, $y$, and $z$. Three binary classifiers consisting of two classes each (i.e., $x \times y$, $x \times z$, and $y \times z$) are used as base learners, and each binary classifier is trained with training images. Each class receives a unique ID as shown in Table 1. To populate the table is straightforward. First, we perform the binary comparison $x \times y$ and tag the class $x$ with the outcome +1, the class $y$ with −1 and set the remaining entries in that column to 0. Thereafter, we repeat the procedure comparing $x \times z$, tag the class $x$ with +1, the class $z$ with −1, and the remaining entries in that column with 0. In the last, we repeat this procedure for binary classifier $y \times z$, and tag the class $y$ with +1, the class $z$ with -1, and set the remaining entries with 0 in that column, where the entry 0 means a "Don't care" value. Finally, each row represents unique ID of that class (e.g., $y = [−1, +1, 0]$). Each binary classifier results a binary response for any input example. Let's say if the outcomes for the binary classifier $x \times y$, $x \times z$, and $y \times z$ are +1, -1, and +1 respectively, then the input example belongs to that class which have the minimum distance from the vector [+1, -1, +1]. So the final answer is given by the minimum distance of

$$\min \text{ dist}\left(\{+1,-1,+1\}, \left(\{+1,+1,0\}, \{-1,0,+1\}, \{0,-1,-1\}\right)\right)$$

In this experiment, we have used Multi-class Support Vector Machine (MSVM) as a set of binary Support Vector Machines (SVMs) for the training and classification.

*Table 1. Unique ID of each class*

|   | $x \times y$ | $x \times z$ | $y \times z$ |
|---|---|---|---|
| $x$ | +1 | +1 | 0 |
| $y$ | -1 | 0 | +1 |
| $z$ | 0 | -1 | -1 |

Figure 3: Sample images from the data set of type (a) apple scab, (b) apple rot, (c) apple blotch, and (d) normal apple

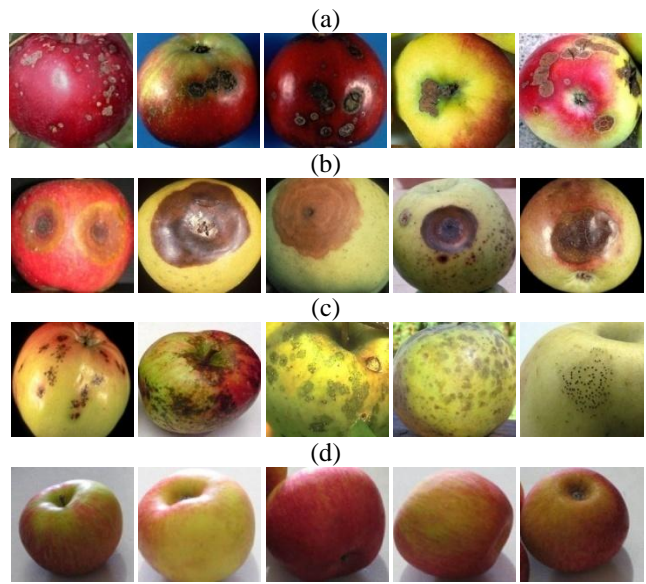

(a)

(b)

(c)

(d)

## RESULTS AND DISCUSSIONS

In this section, first we discuss about the data set of apple fruit diseases and after present a detailed result of the fruit disease identification problem and discuss various issues regarding the performance and efficiency of the system. We consider two color-spaces (i.e. RGB and HSV color-space) and compare the performance of the system under these color spaces.

**Data Set Preparation**

To demonstrate the performance of the proposed approach, we have used a data set of normal and diseased apple fruits, which comprises four different categories: Apple Blotch (104), Apple rot (108), Apple scab (100), and Normal Apple (80). The total number of apple fruit images (N) is 392. Figure 3 depicts the classes of the data set. Presence of a lot of variations in the type and color of apple makes the data set more realistic.

**Result Discussion**

In the quest for finding the best categorization procedure and feature to produce classification, we have analyzed some color, texture and fused image descriptors considering Multiclass Support Vector Machine (MSVM) as a classifier. If we use M images per class for training then remaining N-4*M are used for testing. The accuracy of the proposed approach is defined as,

$$\text{Accuracy}(\%) = \frac{\text{Total number of images correctly classified}}{\text{Total number of images used for testing}} * 100$$

Figure 4 shows the results for different color, texture and fused features. The x-axis represents the number of images per class in the training set and the y-axis represents the average accuracy for the test images. Figure 4(a) depicts the performance of color based descriptors. This experiment shows that CCV performs better than other color based features such as GCH and CDH. One possible explanation is that, GCH uses simply frequency of each color and CDH uses frequency of color difference, however CCV uses frequency of each color in coherent and incoherent regions separately.

Figure 4(b) illustrates the performance of texture based descriptors. CLBP achieves highest accurate recognition rate as comparison to other texture based features such as SEH, LBP and LTP. The basic structures of the SEH are not able to fit in the fruit disease classification problem and shows poor result. The LBP feature uses only the sign information of the local differences, even then, LBP reasonably represent the image local features because sign component preserves the major information of local differences.

Figure 4: Average accuracy (%) considering MSVM as a classifier and using (a) only color feature (i.e. GCH, CCV, CDH), (b) only texture feature (i.e. SEH, LBP, LTP, CLBP), (c) fusion of GCH, LBP (i.e. GCH + LBP) and CCV, LTP (i.e. CCV + LTP), and (d) fusion of CDH, CLBP (i.e. CDH + CLBP), CDH, SEH (i.e. CDH + SEH), and CDH, SEH, CLBP (i.e. CDH + SEH + CLBP).

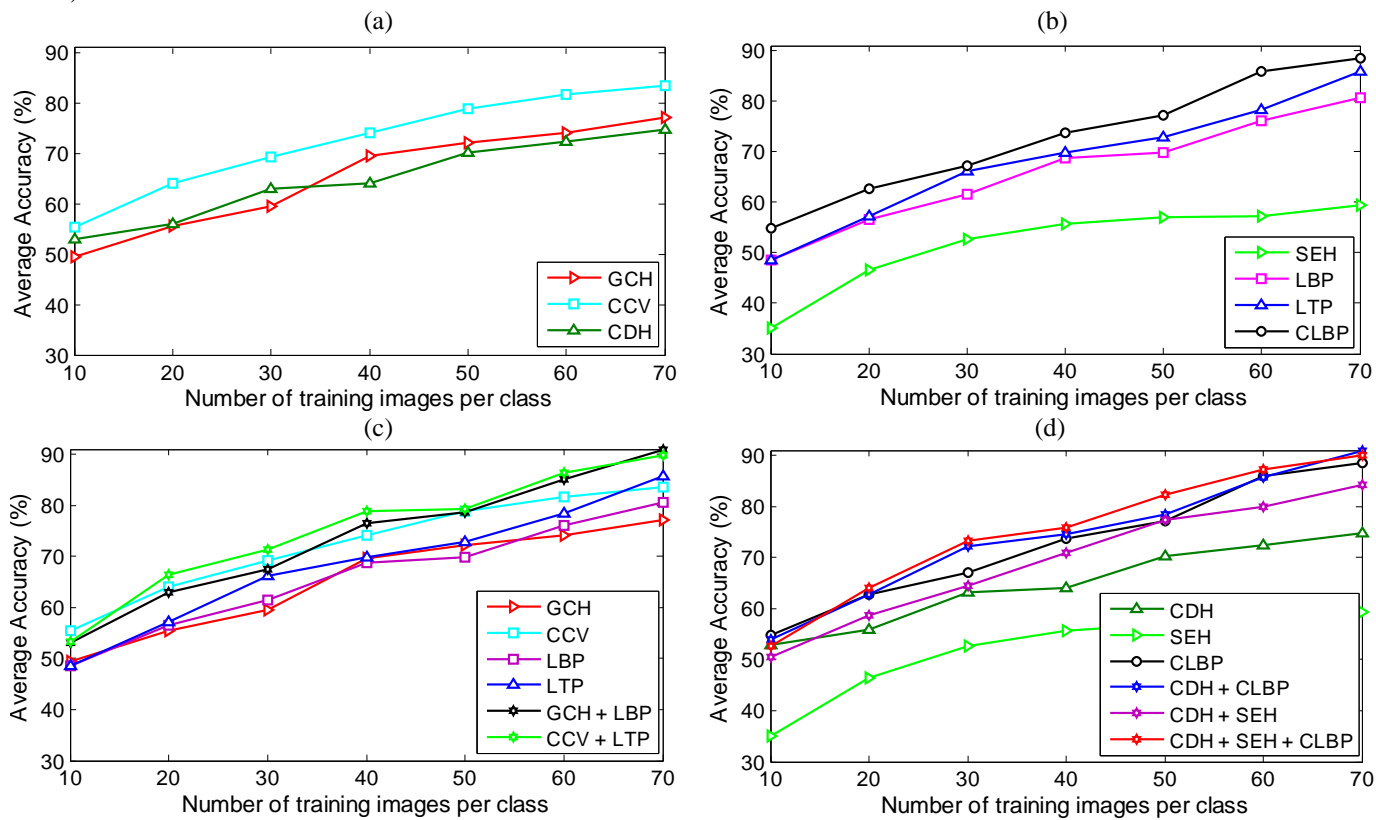

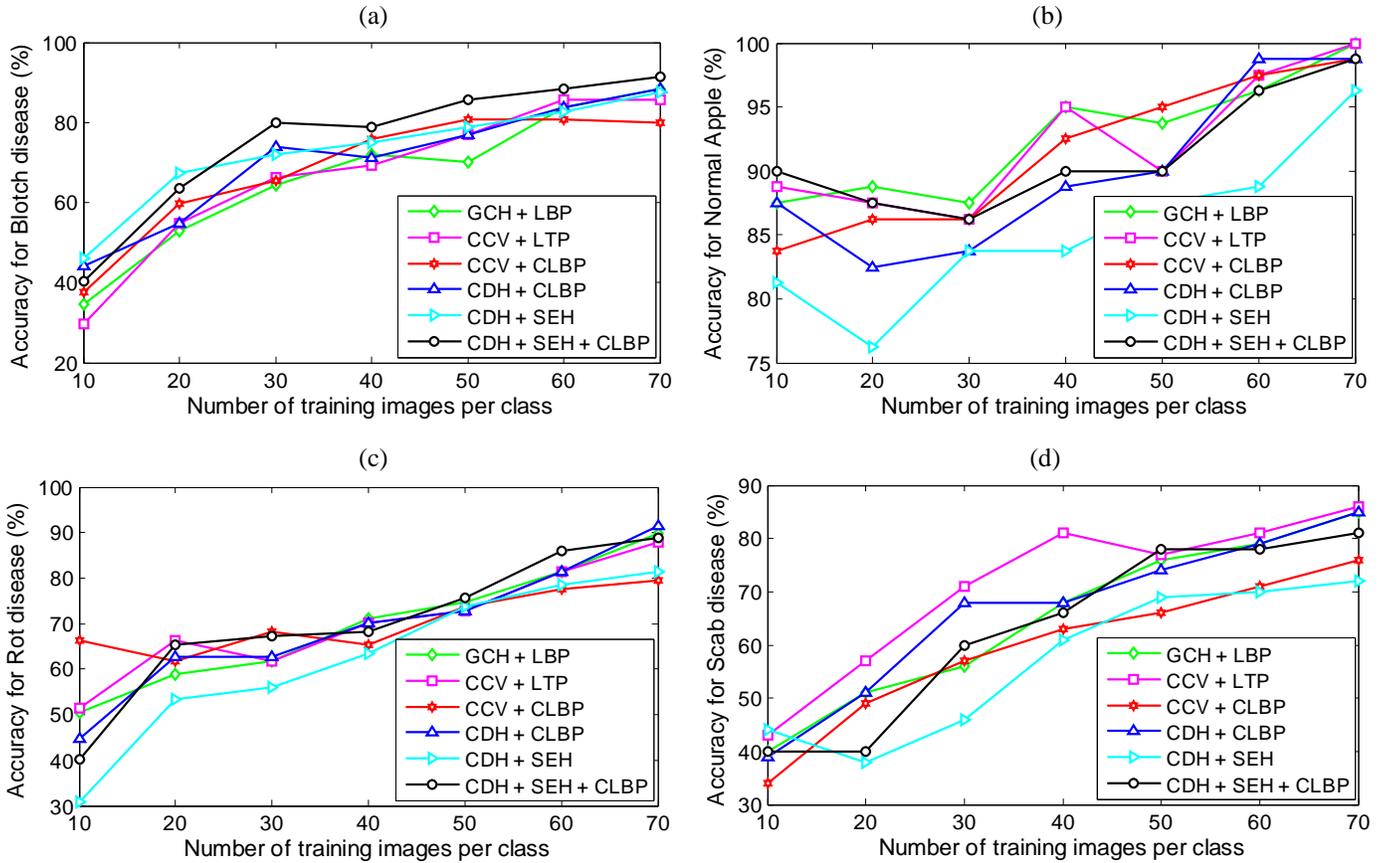

Figure 5: Accuracy using fused features for each type of diseases (a) Blotch, (b) Normal, (c) Rot, and (d) Scab.

The CLBP feature exhibits more accurate result than LBP feature because CLBP feature uses both sign and magnitude component of local differences with original center pixel value. LTP is the extension of LBP and perform better than LBP in this scenario also. Figure 4(c-d) reports the average recognition rate after fusing the color and texture features. We have performed feature level fusion by concatenating one feature after another. The performance of fused features is improved as compared to the stand alone color or texture feature. Fused feature GCH + LBP shows better result than GCH and LBP; CCV + LTP also shows the better result than CCV and LTP, see Figure 4(c).

One interesting combination is CDH + SEH + CLBP, the performance of SEH is poor and the result of CDH is also not too good, but after fusion feature CDH + SEH becomes more discriminative and shows comparable performance. The performance of CDH + SEH is boosted too much after fusion of CLBP with it and the resultant feature (i.e. CDH + SEH + CLBP) performs better than other tested combinations of fusion.

Figure 5(a-d) depicts the performance of each fused features for each category (i.e. blotch, normal, rot and scab respectively). For Blotch and Rot type of diseases, combination CDH + SEH + CLBP is better than other combinations. Normal apples are more accurately recognized by the combination GCH + LBP. Fusion combination CCV + LTP is better suited to detect the Scab type of diseases.

One important aspect when dealing with fruit disease classification is the accuracy per class. This information points out the classes that need more attention when solving the confusions. Figure 6(a-d) depicts the accuracy for each one of 4 classes using GCH + LBP, CCV + LTP, CCV + CLBP, and CDH + SEH + CLBP fused features respectively. Clearly, scab is one class that needs attention as it yields the lowest accuracy when compared to other classes with each feature except CCV + LTP. The accuracy for blotch is better than the rot and scab for each feature except CCV + LTP. Normal Apples are very easily distinguishable with diseased apples and a very good classification result is achieved for the Normal Apples using each features.

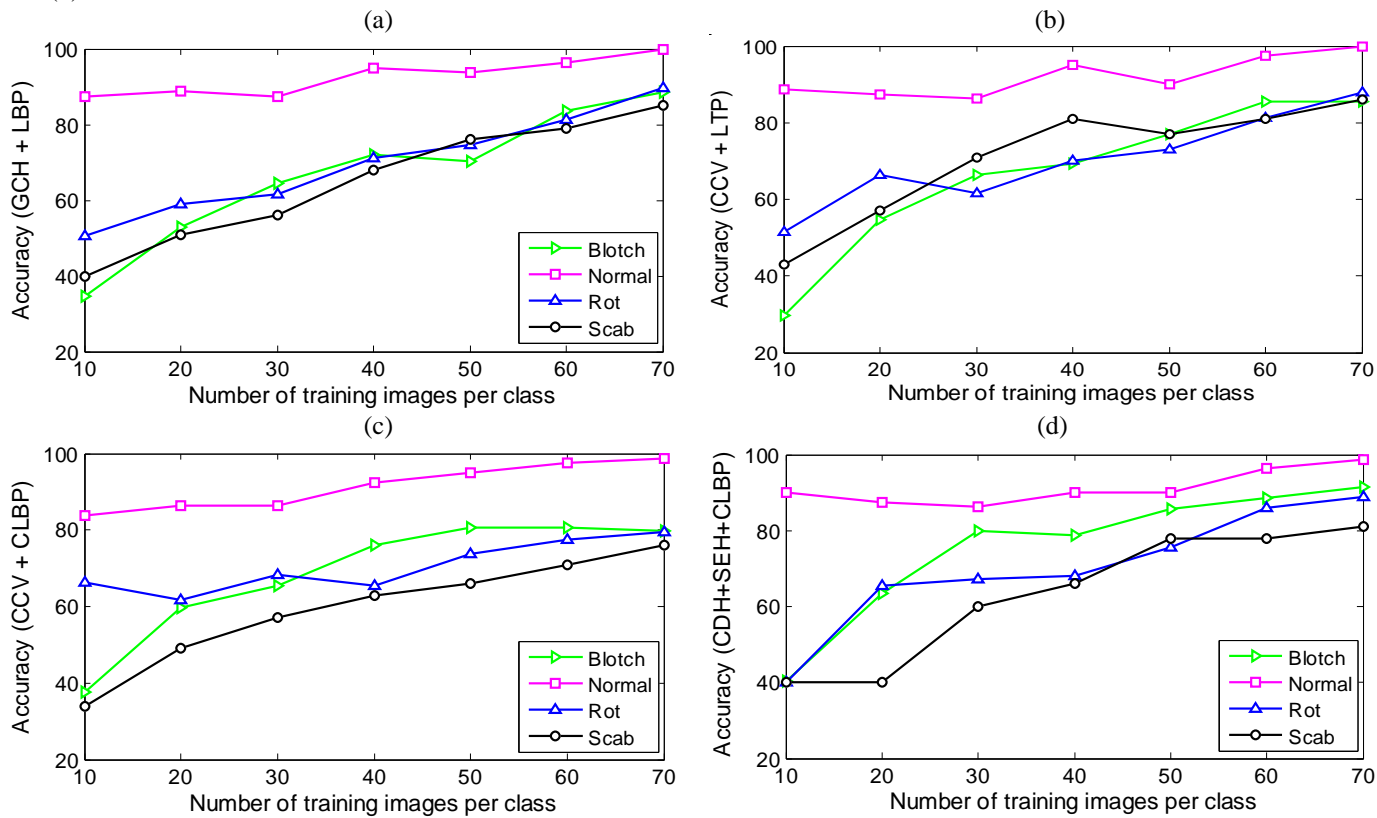

Figure 6: Accuracy per class considering MSVM as a classifier and using (a) GCH + LBP, (b) CCV + LTP, (c) CCV + CLTP, and (d) CDH + SEH + CLBP fused features

Table 2: Fruit disease classification accuracy when MSVM is trained with 70 images per category

| Feature / Category | Blotch | Rot | Scab | Normal | Average |
|---|---|---|---|---|---|
| GCH | 76.92 | 65.42 | 69 | 97.50 | 77.21 |
| CCV | 79.81 | 79.44 | 76 | 98.75 | 83.50 |
| CDH | 78.85 | 63.55 | 63 | 93.75 | 74.79 |
| SEH | 65.38 | 40.19 | 43 | 88.75 | 59.33 |
| LBP | 65.38 | 83.18 | 74 | 100 | 80.64 |
| LTP | 82.70 | 85.98 | 77 | 97.50 | 85.79 |
| CLBP | 83.65 | 87.85 | 84 | 98.75 | 88.56 |
| GCH+LBP | **88.46** | **89.72** | **85** | **100** | **90.80** |
| CCV+LTP | 85.58 | 87.85 | 86 | 100 | 89.86 |
| CCV+CLBP | 79.81 | 79.44 | 76 | 98.75 | 83.50 |
| CDH+CLBP | 88.46 | 91.59 | 85 | 98.75 | 90.95 |
| CDH+SEH | 87.50 | 81.31 | 72 | 96.25 | 84.26 |
| CDH+SEH+CLBP | **91.35** | **88.79** | **81** | **98.75** | **89.97** |

Table 2 depicts the accuracy for each category and also average accuracy by using each feature (i.e. stand alone and also fused). Fused combination achieved GCH+LBP best result when MSVM is trained with 70 images per category. From experiments and results, it is clear those color or texture features are not too distinctive standalone, whereas their performance boosted if they fused with each other.

We also compared proposed method with existing approaches in Table 3 in terms of the application, pre-processing, used features, color space, evaluation criteria and average accuracy. The numbers of clusters in the pre-processing (k-means clustering) are 2 for fruit and vegetable classification problem where it is 4 for the fruit disease classification problems. In this paper the fusion of color and texture features improves the recognition accuracy by nearly 10% as compared to the accuracy of individual color and texture features.

## CONCLUSION

In this paper, an image processing based method is proposed and evaluated for fruit disease classification problem. The introduced method is composed of mainly three steps. In the first step defect segmentation is performed using K-means clustering technique. In the second step color and texture features are extracted and fused with each other. In the third step training and classification are performed on a Multiclass SVM. We used three types of apple diseases namely: Apple Blotch, Apple Rot, and Apple Scab as well as Normal Apples as a case study and evaluated the program.

Table 3: Comparison with existing methods

| Reference | Application | Pre-Processing | Features | Color Space | Training | Evaluation Criteria | Average Accuracy |
|---|---|---|---|---|---|---|---|
| Proposed | Fruit disease classification | K-means with 4 clusters | Color and texture fusion | RGB | Multiclass SVM | Accuracy | ≈ +10% from individual feature |
| Dubey & Jalal (2014a) | Fruit and vegetable classification | K-means with 2 clusters | Color and texture fusion | RGB | Multiclass SVM + KNN | Accuracy and AUC | ≈ +15% from individual feature |
| Dubey & Jalal (2014c) | Fruit disease recognition | K-means with 3 and 4 clusters | ISADH + Gradient filters | HSV | Multiclass SVM + KNN | Accuracy and AUC | >99% |
| Dubey & Jalal (2014b) | Fruit disease classification | K-means with 4 clusters | GCH, CCV, LBP, CLBP | HSV | Multiclass SVM | Accuracy and AUC | 93% |
| Dubey & Jalal (2012b) | Fruit disease classification | K-means with 4 clusters | GCH, CCV, LBP, CLBP | HSV | Multiclass SVM | Accuracy | 93% |
| Dubey & Jalal (2012a) | Fruit and vegetable classification | K-means with 2 clusters | ISADH | HSV | Multiclass SVM | Accuracy | 99% |
| Dubey & Jalal (2013) | Fruit and vegetable species and variety detection | K-means with 2 clusters | GCH, CCV, BIC, Unser, ISADH | HSV | Multiclass SVM + KNN | Accuracy and AUC | 99% |
| Dubey (2012) | Fruit and vegetable classification & Fruit disease classification | K-means with either 2, 3 or 4 clusters | GCH, CCV, BIC, Unser, LBP, CLBP, ISADH | HSV | Multiclass SVM | Accuracy | 99% and 93% |

Our experimental results indicate that the proposed method can significantly support accurate detection and automatic classification of apple fruit diseases. Based on our experiments, we have found that normal apples are easily distinguishable with the diseased apples and fused features shows more accurate result as compared to the standalone features for the classification of apple fruit diseases and achieved satisfactory classification accuracy.